%% file: main.tex
\documentclass[runningheads, envcountsame, a4paper]{llncs}
\usepackage{graphicx}
\usepackage[hyphens]{url}
\usepackage{lineno,hyperref}
\usepackage{amsmath}
\usepackage[utf8]{inputenc} 
\usepackage[T1]{fontenc}    
\usepackage{booktabs}       
\usepackage{amsfonts}       
\usepackage{nicefrac}       
\usepackage{microtype}      
\usepackage{float}
\usepackage{algorithm}
\usepackage[noend]{algpseudocode}
\usepackage{bm}
\usepackage{mathtools}
\usepackage{multicol}
\usepackage{todonotes}
\usepackage{graphicx}
\usepackage{footnote}

\usepackage{lmodern}
\usepackage{tikz}
\usepackage{glossaries}
\usepackage[scientific-notation=true]{siunitx}
\usepackage{subcaption}
\usepackage{common}
\usepackage{url}
\usepackage[misc]{ifsym}
\usepackage{microtype}

\usetikzlibrary{calc}
\usetikzlibrary{positioning}
\usetikzlibrary{chains,fit,shapes}

\newacronym{rnn}{RNN}{Recurrent Neural Network}
\newacronym{cwrnn}{CW-RNN}{Clockwork Recurrent Neural Network}
\newacronym{lmn}{LMN}{Linear Memory Network}
\newacronym[firstplural=Long-Short Term Memory (LSTMs)]{lstm}{LSTM}{Long-Short Term Memory}
\newacronym{gru}{GRU}{Gated Recurrent Unit}
\newacronym{bptt}{BPTT}{Backpropagation Through Time}
\newacronym{cwlmn}{CW-LMN}{Clockwork Linear Memory Network}

\begin{document}
%
\title{Incremental Training of a Recurrent Neural Network Exploiting a Multi-Scale Dynamic Memory}
\titlerunning{Incremental Training of an RNN Exploiting a Multi-Scale Dynamic Memory}
%
%
%
%
%
\author{Antonio Carta{\Letter}\inst{1} \and Alessandro Sperduti\inst{2} \and Davide Bacciu\inst{1}}
\institute{Department of Computer Science, University of Pisa, Pisa, Italy \email{\{antonio.carta,bacciu\}@di.unipi.it} \and
    Department of Mathematics, University of Padova, Padova, Italy \email{sperduti@math.unipd.it}}

\toctitle{Incremental Training of a Recurrent Neural Network Exploiting a Multi-Scale Dynamic Memory}
\tocauthor{Antonio Carta, Alessandro Sperduti, Davide Bacciu}

\maketitle              
\begin{abstract}
    The effectiveness of recurrent neural networks can be largely influenced by their ability to store into their dynamical memory information extracted from input sequences at different frequencies and timescales. Such a feature can be introduced into a neural architecture by an appropriate modularization of the dynamic memory. In this paper we propose a novel incrementally trained recurrent architecture targeting explicitly multi-scale learning. First, we show how to extend the architecture of a simple RNN by separating its hidden state into different modules, each subsampling the network hidden activations at different frequencies. Then, we discuss a training algorithm where new modules are iteratively added to the model to learn progressively longer dependencies. Each new module works at a slower frequency than the previous ones and it is initialized to encode the subsampled sequence of hidden activations. Experimental results on synthetic and real-world datasets on speech recognition and handwritten characters show that the modular architecture and the incremental training algorithm improve the ability of recurrent neural networks to capture long-term dependencies.
\end{abstract}

\input{body.tex}

%
%
\bibliographystyle{splncs04}
\bibliography{biblio}
\end{document}

%% file: body.tex
\section{Introduction}
Time series, such as speech and music sound waves~\cite{gravesSPEECHRECOGNITIONDEEP2013,oordWaveNetGenerativeModel2016}, and raw sensor data from several domains are sampled at high frequencies, generating large datasets of long and fast-flowing sequences. Recurrent neural networks must dynamically extract information from these samples at different frequencies. Unfortunately, these data present two challenges that limit the application of recurrent neural networks. First, vanilla RNN~\cite{elmanFindingStructureTime1990} and LSTM~\cite{hochreiterLongShortTermMemory1997} do not take into account the importance of different frequencies when processing a sequence, making it difficult to capture high and low frequencies together. Furthermore, low-frequency information requires capturing long-term temporal dependencies in the sequence, which are difficult to learn by stochastic gradient descent (SGD) due to the vanishing gradient problem~\cite{hochreiterLongShortTermMemory1997,pascanuDifficultyTrainingRecurrent2013}. The most popular solutions to address this problem is to subsample the sequence, discarding useful data, or to process the sequence to extract hardcoded features that reduce the sampling rate. Both solutions circumvent the problem by avoiding to learn features for short-term dependencies while shortening the long-term dependencies. Both approaches simplify the learning problem but have the drawback that useful information is discarded when it could be used to improve the quality of network predictions.

In this paper, we propose a new model, called the MultiScale LMN (MS-LMN), a recurrent neural network designed to easily represent and learn short-term and long-term dependencies. The architecture of the MS-LMN partitions the memory state of the model into separate modules, and each module is updated with a different frequency. Therefore, high frequency and low frequency information is separated such that the short-term dependencies do not interfere with the long-term ones.

To improve the learning of long-term dependencies we propose an incremental training algorithm. Since short-term dependencies are easier to learn, the model can be trained to learn them before the long-term ones, by learning only the parameters relevant to the high-frequency memory state. Incrementally, new modules can be added to the MS-LMN to learn longer dependencies. Each module is initialized to memorize the subsampled sequence and finetuned with SGD. The complexity of the model grows according to the complexity of the learning task, which is given by the maximum dependency length.

Synthetic and real world time series datasets are used to assess the proposed model. In particular, we present empirical evidence that the model is able to generate a long sequence, learning to model both short-term and long-term dependencies in the data. Moreover, experiments on speech recognition and handwritten character recognition show that the proposed architecture and associated incremental training algorithm can effectively learn on real-world datasets when using high-frequency features.

In summary, the main contributions of the paper are as follows:
\begin{itemize}
    \item the proposal of a novel hierarchical RNN architecture designed to model low frequency and high frequency features separately;
    \item the proposal of a novel incremental training algorithm for RNNs designed to incrementally learn long-term dependencies;
    \item an experimental assessment of the proposed approach that shows the benefits of incremental training on synthetic and real-world datasets.
\end{itemize}

\section{Background}
In this section we describe two models, the Linear Memory network (LMN)~\cite{bacciuLinearMemoryNetworks2019}, and the linear autoencoder for sequences (LAES) \cite{sperdutiExactSolutionsRecursive2006}. The proposed architecture extends the LMN with a modular design. The LAES training algorithm is a fundamental step of the incremental training.

\subsection{Linear Memory Network}
The dynamic multiscale architecture put forward in this paper is built on the Linear Memory Network (LMN)~\cite{bacciuLinearMemoryNetworks2019} concepts.
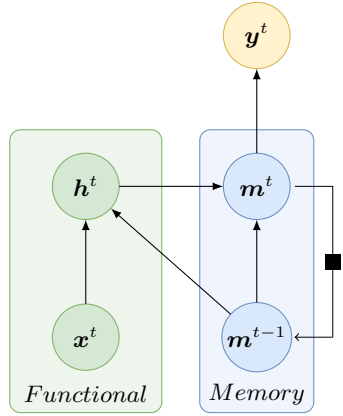
\begin{figure}[t]
    \centering
    \input{img/lmn.tex}
    \caption{A Linear Memory Network, where $\vx^t, \vh^t, \vm^t, \vy^t$ are the input, hidden, memory and output vector, respectively. The edge with the small black square represents a time delay.} \label{fig:lmn}
\end{figure}
The LMN is a recurrent neural network that updates its memory state using a nonlinear mapping of the input and a linear mapping of its memory state.  The LMN computes a hidden state $\vh^t$ and a memory state $\vm^t$ as follows:
\LMNeq
\noindent where $\mW^{xh} \in \mathbb{R}^{N_h \times N_x}$, \mbox{$\mW^{mh} \in \mathbb{R}^{N_h \times N_m}$}, $\mW^{hm} \in \mathbb{R}^{N_m \times N_h}$, $\mW^{mm} \in \mathbb{R}^{N_m \times N_m}$ are the model parameters matrices, with $N_x,\ N_h,\ N_m$ the input size, hidden size and memory size respectively, and $\sigma$ is a non-linear activation function ($tanh$ for the purpose of this paper). The memory state $\vm^t$ is the output of the entire layer, which is passed to the next layer. The output $\vy^t$, is computed by a linear layer:
\begin{equation}\label{eq:lmn_readout}
    \vy^t = \mW^{my}\vm^t.
\end{equation}
A schematic view of the LMN is shown in Fig.~\ref{fig:lmn}. Notice that the linearity of the recurrence does not limit the expressive power of the entire model since given an RNN such that $\vh^t = \sigma(\tilde{\mW}^{xh} \vx^t + \tilde{\mW}^{hh} \vh^{t-1})$, it is possible to initialize an equivalent LMN that computes the same hidden activations by setting \mbox{$\mW^{xh} = \tilde{\mW}^{xh}$}, $\mW^{mh} = \tilde{\mW}^{hh}$, $\mW^{hm} = \mathbb{I}$, $\mW^{mm} = 0$. The linearity of the memory update plays a key role during the initialization of a new module in the incremental training algorithm.

\subsection{Linear Autoencoder for Sequences} \label{sec:laes}
The linearity of the LMN memory update provides an opportunity to explicitly optimize the encoding learned by the model. Ideally, we would like to train the model to encode the entire sequence to avoid forgetting past elements. However, this is an autoencoding problem with long-term dependencies, which is difficult to solve by SGD. Fortunately, we can exploit the linearity to find the optimal autoencoder with a closed form solution with the linear autoencoder for sequences (LAES) \cite{sperdutiExactSolutionsRecursive2006}. LAES is a recurrent linear model that is able to memorize an input sequence by encoding it into a hidden memory state vector recursively updated which represents the entire sequence. Given the memory state, the original sequence can be reconstructed.

Given a sequence $s = \vx^1, \hdots, \vx^l$, where $\vx^i \in \mathbb{R}^a$, a linear autoencoder computes the memory state vector $\vm^t \in \mathbb{R}^p$, i.e. the encoding of the input sequence up to time $t$, using the following equations:
\begin{align}
    \vm^t = \mA\vx^t + \mB\vm^{t-1},  \label{eq:ae_enc} \\
    \begin{bmatrix} \vx^t \\ \vm^{t-1} \end{bmatrix} = \mC\vm^t, \label{eq:ae_dec}
\end{align}
\noindent where $p$ is the memory state size, $\mA \in \mathbb{R}^{p \times a}$, $\mB \in \mathbb{R}^{p \times p}$ and $\mC \in \mathbb{R}^{(a+p) \times p}$ are the model parameters. Eq.~(\ref{eq:ae_enc}) describes the encoding operation, while Eq.~(\ref{eq:ae_dec}) describes the decoding operation.

\subsubsection{Training Algorithm}
The linearity of the LAES allows us to derive the optimal solution with a closed-form equation, as shown in \cite{sperdutiEfficientComputationRecursive2007}. For simplicity, let us assume that the training set consists of a single sequence $\{\vx_1, \hdots, \vx_l\}$ and define $\mM \in \mathbb{R}^{l \times p}$ as the matrix obtained by stacking by rows the memory state vectors of the LAES at each timestep. From Eq.~(\ref{eq:ae_enc}) it follows that:

\begin{equation}
\underbrace{
\begin{bmatrix}
{\vm^1}^\top \\ {\vm^2}^\top \\ {\vm^3}^\top \\ \vdots \\ {\vm^l}^\top
\end{bmatrix}}_{\mM} =
\underbrace{\begin{bmatrix}
{\vx^1}^\top & 0 & \hdots & 0 \\
{\vx^2}^\top & {\vx^1}^\top & \hdots & 0 \\
\vdots & \vdots & \ddots & \vdots \\
{\vx^l}^\top & {\vx^{l-1}}^\top & \hdots & {\vx^1}^\top
\end{bmatrix}}_{\mXi}
\underbrace{\begin{bmatrix}
\mA^\top \\ \mA^\top \mB^\top \\ \vdots \\ \mA^\top {\mB^{l-1}}^\top
\end{bmatrix}}_{\mOmega}.
\end{equation}

The matrix $\mXi \in \mathbb{R}^{l \times la}$ contains the reversed subsequences of $s$, while \mbox{$\mOmega \in \mathbb{R}^{la \times p}$} contains the matrices to encode the input elements for up to $l$ timesteps.
The encoder parameters $\mA$ and $\mB$ can be identified by exploiting the singular value decomposition (SVD) $\mXi = \mV \mSigma \mU^\top$, where imposing $\mU^\top \mOmega = \mI$ yields $\mOmega = \mU$. Given this additional constraint, we can then exploit the structure of $\mXi$ to recover $\mA$, $\mB$, and the decoder parameters $\mC = \begin{bmatrix} \mA^\top \\ \mB^\top \end{bmatrix}$. Specifically, $\mOmega=\mU$ is satisfied by using the matrices
\begin{equation*}
 \mP  \equiv  \left[\begin{array}{c} \mI_{a} \\ \mzero_{a(l - 1)\times a}\end{array}\right],\ \ \mbox{and}\ \ \mR  \equiv  \left[\begin{array}{ll} \mzero_{a\times a(l - 1)} &  \mzero_{a\times a}\\ \mI_{a(l - 1)} & \mzero_{a(l - 1)\times a}\end{array}\right],
\end{equation*}
to define
$\mA\equiv \mU^\top \mP$ and \mbox{$\mB\equiv \mU^\top  \mR \mU$}, where $\mI_{u}$ is the identity matrix of size $u$, and $\mathbf{0}_{u\times v}$ is the zero matrix of size $u\times v$.
The algorithm can be easily generalized to multiple sequences by stacking the data matrix $\mXi_q$ for each sequence $s^q$ and padding with zeros to match sequences length, as shown in \cite{sperdutiExactSolutionsRecursive2006}.

The optimal solution reported above allows encoding the entire sequence without errors with a minimal number of memory units $p = rank(\mXi)$. Since we fix the number of memory units before computing the LAES, we approximate the optimal solution using the truncated SVD, which introduces some errors during the decoding process.
The computational cost of the training algorithm is dominated by the cost of the truncated SVD decomposition, which for a matrix of size $n \times m$ is $\mathcal{O}(n^2 m)$. Given a dataset of sequences with lengths $l^1, \hdots, l^{S}$, with $l=\sum_{i=1}^S l^i$, $l_{max} = \max_i \{l^i\}$, we have $\mXi \in \mathbb{R}^{l \times l_{max}a}$, which requires $\mathcal{O}(l_{max} l a)$ memory for a dense representation. The memory usage can be reduced for sparse inputs, such as music in a piano roll representation, by using a sparse representation. To mitigate this problem in a more general setting, we approximate the SVD decomposition using the approach proposed in \cite{pasaPretrainingRecurrentNeural2014}, which computes the SVD decomposition with an iterative algorithm that decomposes $\mXi$ in slices of size $l \times a$ and requires $\mathcal{O}(l a)$ memory.

\section{A Dynamic Multiscale Memory for Sequential Data} \label{sec:ms_lmn}
\begin{figure}[t]
    \centering
    \input{img/ms_lmn.tex}
    \caption{Architecture of the MultiScale Linear Memory Network with $3$ memory modules. Dashed connections are affected by the subsampling, while connections with small black squares represent time delays.} \label{fig:ms_lmn}
\end{figure}
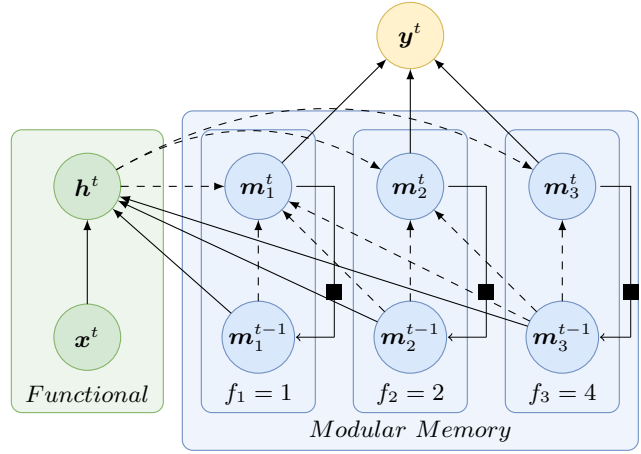

Learning long-term dependencies from high-frequency data with recurrent models is a difficult problem to solve with classic architectures. Due to the vanishing gradient, short-term dependencies will dominate the weights update. Popular solutions like the LSTM alleviate this problem in some practical scenarios but do not solve it completely. To address this problem, we extend the LMN with a modular memorization component, divided into separate modules, each one responsible to process the hidden state sequence at different timescales, as shown in Fig.~\ref{fig:ms_lmn}. The modules responsible for longer timescales subsample the hidden states sequence to focus on long-term interactions in the data and ignore the short-term ones. In practice, we separate the memory state into $g$ different substates, each one updated with exponentially longer sampling rates $1, 2, ..., 2^g$. The connections affected by the subsampling are shown with dashed edges in Fig.~\ref{fig:ms_lmn}. The number of modules can be chosen given the maximum length of the sequences in the training set $l_{max}$. Given $l_{max}$, the maximum number of different frequencies is $g = \lfloor \log{l_{max}} \rfloor$, which means the model only needs a logarithmic number of modules. Each memory module is connected only to slower modules, and not vice-versa, to avoid interference of the faster modules with the slower ones. The organization of the memory into separate modules with a different sampling rate is inspired by the Clockwork RNN\cite{koutnikClockworkRNN2014}, which is an RNN with groups of hidden units that work with different sampling frequencies. Differently from the Clockwork RNN, we apply this decomposition only to the memory state. Furthermore, by adopting a linear recurrence we can achieve better memorization of long sequences, as we will see in the experimental results in Section \ref{sec:experiments}.

The model update computes at each timestep $t$ an hidden state $\vh^t$ and $g$ memory states $\vm_1^t, \hdots, \vm_g^t$ as follows:

\begin{eqnarray}
 \vh^t &=& \sigma(\mW^{xh} \vx^t + \sum_{i=1}^{g} \mW^{m_i h} \vm^{t-1}_i), \label{eq:ms_lmn_h} \\
 \vm^t_k &=& \begin{cases}
 \vm_{new}^t & if\ t\mod{2^{k-1}} = 0\  \\
 \vm_k^{t-1} & otherwise
 \end{cases} \ \ \ \ \forall k \in 1, \hdots, g, \label{eq:ms_lmn_m} \\
 \vm_{new}^t &=& \mW^{h m_k} \vh^t + \sum_{i=k}^{g} \mW^{m_i m_k} \vm^{t-1}_i,
\end{eqnarray}
where $\vx^t \in \mathbb{R}^{N_x}$, $\vh^t \in \mathbb{R}^{N_h}$, $\vm_k^t \in \mathbb{R}^{N_m}$. The subsampling of the hidden state sequence is performed by choosing when to update the memory state using the modulo operation. The network output can be computed from the memory modules' output as follows:
\begin{equation} \label{eq:ms_lmn_out}
    \vy^t = \sum_{i=1}^g \mW^{m_i y} \vm_i^t.
\end{equation}

Fig.~\ref{fig:ms_lmn} shows a schematic view of the architecture.
For a more efficient implementation, more amenable to parallel architectures like GPUs, we can combine all the operations performed by Eqs.~(\ref{eq:ms_lmn_h})-(\ref{eq:ms_lmn_out}) for each module into a single matrix multiplication. In the following, we show the procedure for Eq.~(\ref{eq:ms_lmn_m}) since the same approach can be applied to Eqs.~(\ref{eq:ms_lmn_h}) and (\ref{eq:ms_lmn_out}). First, we notice that the memory modules are ordered by frequency, from fastest to slowest, and their sampling frequencies are powers of $2$. As a consequence, if module $i$ is active, then all the modules $j$ with $j < i$ are also active since $t \mod 2^i = 0$ implies $t \mod 2^j = 0$ whenever $j \geq 0$ and $j < i$. Therefore, we only need to find the maximum index of the active modules $i_{max}^t = max \{ i\ |\ t \mod 2^{i-1} = 0\ \land\ i \leq g \}$ to know which memory modules must be updated. We can combine the activations and parameters of the modules together as follows:
\begin{align*}
    \vm^t = \begin{bmatrix} {\vm_1^t}^\top \\ \vdots \\ {\vm_k^t}^\top \end{bmatrix}, \mW^{hm} = \begin{bmatrix} \mW^{h_1 m} \\ \vdots \\ \mW^{h_g m} \end{bmatrix}, \\
    \mW^{mm} = \begin{bmatrix}
        \mW^{m_1 m_1} & & \hdots & \mW^{m_g m_1} \\
        0 & \mW^{m_2 m_2} & \hdots & \mW^{m_g m_2} \\
        \vdots & \ddots & \ddots & \vdots \\
        0 & \hdots & 0 & \mW^{m_g m_g}
    \end{bmatrix}.
\end{align*}
\input{img/mslmn_blocks.tex}
Eq.~(\ref{eq:ms_lmn_m}) becomes:
\begin{align}
    \vm^t[:i_{max}] &= \mW^{hm}[:i_{max}] \vh^t[:i_{max}] + \mW^{mm}[:i_{max}]\vm^t[:i_{max}], \label{eq:mslmn_m_block1}\\
    \vm^t[i_{max}:] &= \vm^{t-1}[i_{max}:], \label{eq:mslmn_m_block2}
\end{align}
where the slicing operator $\vm[i_{start}:i_{end}]$ returns a vector $\vm[i_{start}:i_{end}] \in \mathbb{R}^{i_{end} - i_{start}}$ with $\vm[i_{start}:i_{end}]_{k} = \vm[i_{start} + k]$, which is used to select the vector of the currently active memory modules.
Using Eqs.~(\ref{eq:mslmn_m_block1})-(\ref{eq:mslmn_m_block2}), the subsampling is performed by finding $i_{max}$ which determines which slice of $\vm^t$ must be updated or remain constant. Notice that $\mW_{mm}$ is a block diagonal matrix, therefore it has fewer parameters than a corresponding LMN with $gN_m$ hidden units. Fig.~\ref{fig:cwrnn-blocks} shows the block structure of the MS-LMN parameters for the memory state update, with the darker blocks being the currently active modules.
\section{Incremental Training of the Multiscale Memory}\label{sec:mslmn_pretrain}
The MS-LMN is designed to model long-term dependencies at different frequencies. To better exploit the modular architecture, we propose a constructive algorithm that incrementally trains the network by adding a new module at each step. During an incremental step the algorithm performs the following operations:
\begin{enumerate}
    \item Add a new module with a lower sampling rate.
    \item Initialize the new module to encode the hidden state into the memory state.
    \item Finetune the model by SGD.
\end{enumerate} 
Incremental training helps to gradually learn longer dependencies, while the initialization ensures that the new module is able to encode a long sequence of activations, which helps to extract long-term dependencies from the data.
In this section we show how to train the MS-LMN by describing the incremental construction of the model and the initialization procedure of the new module.
\begin{figure}[t]
    \centering
    \input{img/ms_lmn_init.tex}
    \caption{MS-LMN pre-training: {\it (right side of picture)} the $g+1$th new memory model is added. New connections are shown in red.  The LAES training algorithm is used to pre-train the weight matrices $\mA_{g+1}$, $\mB_{g+1}$, and $\mW_{g+1}^{out}$, while the remaining new connections are initialized with zeros.}
    \label{fig:mslmn_pretrain}
\end{figure}
\subsection{MS-LMN Incremental Training}
The training algorithm for the MS-LMN works incrementally. During the first iteration, a MS-LMN with a single module is trained by SGD. After a fixed number of epochs\footnote{This is a simple but suboptimal strategy that has been adopted to show the effectiveness of the proposed model. This strategy can easily be improved so to optimize the performance on the validation set.}, an additional memory module with a slower frequency is added,  and it is initialized to encode the hidden activations of the current model using the LAES training algorithm (see Fig.~\ref{fig:mslmn_pretrain}). The resulting model is trained with SGD and the process is repeated until all the modules have been added.

The addition of a new module works as follows. Let us assume to have already trained an MS-LMN with $g$ modules as defined in Eqs.~(\ref{eq:ms_lmn_h})-(\ref{eq:ms_lmn_m}). We collect the sequences of hidden activations $\vh^1, \hdots, \vh^t$ computed by the model for each sample in the training set, subsample them with frequency $2^g$, which is the subsampling rate of the new module, and train a LAES with parameters $\mA_{g+1}$ and $\mB_{g+1}$. A new module with sampling frequency $2^{g}$ is added to the model and the new connections are initialized as follows:
\begin{equation*}
 \mW^{h m_{g+1}} = \mA_{g+1}, \ \ \ \mW^{m_{g+1} m_{g+1}} = \mB_{g+1}, \ \ \ \mW^{m_{g+1} o} = \mW_{g+1}^{out},
\end{equation*}
where $\mW_{g+1}^{out}$ is obtained by training a linear model (via pseudoinverse) to predict the desired output $\vy^t$ from the memory state of the entire memory $\vm_1^t, \hdots, \vm_{g+1}^t$. The remaining new parameters are initialized with zeros, while the old parameters remain unchanged. After the initialization of the new connections, the entire model is trained end-to-end by gradient descent. Notice that during this phase the entire model is trained, which means that the connections of the old modules can still be updated. This process is repeated until all the memory modules have been added and the entire model has been trained for the last time. Algorithm \ref{alg:ms_lmn_train} shows the pseudocode for the entire training procedure.
\input{pcode/pseudo_train_ms_lmn.tex}
The computational cost of the incremental training process is the sum of the cost of training the autoencoders and the cost of the SGD. In practice, despite the worse computational cost, training the LAES requires a negligible amount of time compared to the SGD, due to the large number of iterations needed by the latter to reach convergence. It is important to notice that incremental training is less expensive than training the full model with all memory modules from scratch by SGD since during the first phases the model has a lower number of modules, and therefore it has a lower computational cost.

\section{Experimental Analysis} \label{sec:experiments}
In the following experiments, we study the ability of the modular memorization component of the MS-LMN to capture long-term dependencies in three different sequential problems. The model is compared against LSTM, LMN, and CW-RNN, including all datasets used in the original CW-RNN publication \cite{koutnikClockworkRNN2014}.  Each model is a single layer recurrent neural network with \emph{tanh} activation function (only in the functional component for LMN and MS-LMN). We tested the MS-LMN on audio signals since they are a natural choice to assess the modular memory, which works by subsampling its input sequence. The first task, \emph{Sequence Generation}, is a synthetic problem that requires the model to output a given signal without any eternal input. The second task, \emph{Spoken Word Classification}, is a sequence classification task that uses a subset of spoken words extracted from TIMIT~\cite{garofoloDARPATIMITAcousticphonetic1993}. The task is designed to have long-term dependencies by only considering a restricted subset of words that have a common suffix. The third experiment performs handwritten character recognition on the IAM-OnDB online dataset~\cite{UnconstrainedOnlineHandwriting}. Throughout all the experiments we use the Adam optimizer~\cite{kingmaAdamMethodStochastic2014} with L2 weight decay. The source code is available online\footnote{\url{https://github.com/AntonioCarta/mslmn}}.

\subsection{Sequence Generation} \label{sec:wavegen}
The Sequence Generation task is a synthetic problem that requires the model to output a given signal without any eternal input. We extracted a sequence of 300 data points generated from a portion of a music file, sampled at 44.1 kHz starting from a random position. Sequence elements are scaled in the range \( [-1,1] \). The task stresses the ability to learn long-term dependencies by requiring the model to encode the entire sequence without any external input. Notice that this task is used only to assess the effectiveness of the modular architecture and therefore we do not use the incremental pretraining.
The CW-RNN\cite{koutnikClockworkRNN2014} is the state-of-the-art on this task, where it reaches better results than comparable RNNs and LSTMs.
We tested each model with $4$ different numbers of parameters in $\{100, 250, 500, 1,000\}$, by varying the number of hidden neurons. The models are trained to optimize the Normalized MSE. Table~\ref{tab:num-params} shows the most relevant hyperparameters of the best configuration for each model which were found with a random search. 
\begin{table}[t]
    \centering
    \small
    \caption{Number of parameters and corresponding number of hidden units for different configurations trained on the Sequence Generation task. For CW-RNN and MS-LMN we show the total number of units and the number of units for each module (the latter between parenthesis).\label{tab:num-params}}
    \begin{tabular}{cccccc}
        \toprule
        \#parameters & RNN   & LSTM  & CW-RNN     & LMN        & MS-LMN \\
                     & $N_h$ & $N_h$ & $N_h, (\frac{N_h}{g})$ & $N_h, N_m$ & $N_h, N_m, (\frac{N_m}{g})$ \\
        \midrule
        100  & 9  & 4  & 9, (1)  & 4, 6   & 1, 9, (1) \\
        250  & 15 & 7  & 18, (2) & 7, 10  & 1, 18, (2) \\
        500  & 22 & 10 & 27, (3) & 11, 13 & 1, 27, (3) \\
        1,000 & 31 & 15 & 36, (4) & 2, 29 & 1, 36, (4) \\
        \bottomrule
    \end{tabular}
\end{table}
The CW-RNN and MS-LMN use $9$ modules. Notice that adding more modules would be useless since the sequence length is $300$. The number of hidden units for the CW-RNN and of memory units for the MS-LMN is the number of modules times the number of units per module. For the LSTM, we obtained the best results by initializing the forget gate to $5$, as suggested in \cite{cumminsLearningForgetContinual2000}.

\begin{table}[t]
    \centering
    \small
    \caption{Results for the Sequence Generation task. Performance computed using NMSE (lower is better).\label{tab:gen-hyperparams}}
    \begin{tabular}{lccccc}
        \toprule
                            & RNN                    & LSTM          & CW-RNN        & LMN                    & MS-LMN \\
        \midrule
        Hidden Units        & 31                     & 15            & 36            & 2                      & 1 \\
        Memory Units        & /                      & /             & /             & 29                     & 36 \\
        Learning Rate       & \num{1e-03}            & \num{1e-02}   & \num{5e-05}   & \num{5e-04}            & \num{5e-03} \\
        \# parameters       & 1,000                   & 1,000          & 1,000          & 1,000                   & 1,000 \\
        Epochs              & 6,000                   & 12,000         & 2,000          & 5,000                   & 8,000 \\
        \midrule
        NMSE ($10^{-3}$)                & 79.5                & 20.7       & 12.5       & 38.4                & \textbf{0.116} \\
        \bottomrule
    \end{tabular}
\end{table}

\begin{figure}
    \centering
    \input{img/gen_seq.tex}
    \caption{Generated output for the Sequence Generation task. The original sequence is shown with a dashed blue line while the generated sequence is a solid green line.}\label{fig:gen-plots}
\end{figure}

The results of the experiments are reported in Table~\ref{tab:gen-hyperparams}. Fig.~\ref{fig:gen-plots} shows the reconstructed sequence for each model. The results confirm those found in \cite{koutnikClockworkRNN2014} for the CW-RNN and show a clear advantage of the CW-RNN over the RNN and LSTM. The LMN provides an approximation of the sequence, slightly worse than that of the LSTM, which closely follows the global trend of the sequence, but it is not able to model small local variations. The MS-LMN obtains the best results and closely approximates the original sequence. The model combines the advantages of a linear memory, as shown by the LMN performance, with the hierarchical structure of the CW-RNN. The result is that the MS-LMN is able to learn to reproduce both the short-term and long-term variations of the signal.

\subsection{Common Suffix TIMIT} \label{sec:timit}
TIMIT~\cite{garofoloDARPATIMITAcousticphonetic1993} is a speech corpus for training acoustic-phonetic models and automatic speech recognition systems. The dataset contains recordings from different speakers in major American dialects, and it provides a train-test split and information about words and phonemes for each recorded audio file.

Since we are interested in the ability of the assessed models to capture long-term dependencies, we extract from TIMIT a subset of words that have a common suffix, following the preprocessing of the original CW-RNN paper \cite{koutnikClockworkRNN2014}. We took $25$ words pronounced by $7$ different speakers, for a total of $175$ examples. The chosen words can be categorized into $5$ clusters based on their suffix:
\begin{itemize}
    \item \emph{Class 1}: making, walking, cooking, looking, working;
    \item \emph{Class 2}: biblical, cyclical, technical, classical, critical;
    \item \emph{Class 3}: tradition, addition, audition, recognition, competition;
    \item \emph{Class 4}: musicians, discussions, regulations, accusations, conditions;
    \item \emph{Class 5}: subway, leeway, freeway, highway, hallway.
\end{itemize}
The common suffix makes the task more difficult since the model is forced to remember the initial part of the sequence to correctly classify each word.

Each file in the TIMIT dataset is a WAV containing the recording of a sentence, therefore we trimmed it to select a single word using the segmentation metadata provided with the dataset. The words extracted from this procedure are preprocessed to extract a sequence of MFCC coefficients using a window length of 25ms, a step of 1ms, preemphasis filter of 0.97, 13 cepstrum coefficients, where we replaced the zeroth cepstral coefficient with the \emph{energy} (the log of total frame energy). As a result, we obtain $13$ features for each timestep. We normalized each feature to have mean $0$ and variance $1$.

To allow a direct comparison with the work in~\cite{koutnikClockworkRNN2014}, we split the dataset by taking 5 words for training and 2 for test from each class. This split ensures a balanced train and test set. During training we added gaussian noise to the sequence with a standard deviation of $0.6$. Due to the small size of the dataset, we do not use an additional train-validation split and we use the clean version of the training set as a validation set, as done in \cite{koutnikClockworkRNN2014}.
Unfortunately, \cite{koutnikClockworkRNN2014} does not provide the exact split used in their experiments. Given the small size of the dataset, in our experiments we have found a great variance between different splits. Therefore, we cannot directly compare against the results in \cite{koutnikClockworkRNN2014} and we decided to train the CW-RNN with our train-test split. To ensure the reproducibility of our work and a fair comparison for future work, we provide the splits in the supplementary material.

We found the best hyperparameters with a random search on the batch size in $\{1, 25\}$, l2 decay in $\{0, 10^{-3}, 10^{-4}, 10^{-5}\}$, learning rate in $\{10^{-3}, 10^{-4}\}$ and hidden units per module in $[5, 40]$. Each model is trained to minimize the cross-entropy loss. When using a batch size equal to \( 25 \) we keep the classes balanced by taking one sample per class. The CW-RNN and MS-LMN use 7 modules, since the longer sequence has 97 data points. We initialize the LSTM forget gate to $5$, as suggested in \cite{cumminsLearningForgetContinual2000}.

Table \ref{tbl:timit_results} shows the results of the experiment. The CW-RNN and MS-LMN obtain much better results than the LMN due to their ability to learn long-term dependencies. The MS-LMN shows superior results compared to the CW-RNN, and the incremental training improves the performance of the LMN substantially.
\begin{table}[b]
    \centering
    \caption{Test set accuracy on Common Suffix TIMIT. Variance computed by training with different random seeds for $5$ times using the best hyperparameters found during the model selection.}\label{tbl:timit_results}
    \begin{tabular}{llll}
    \toprule
    Model                    &$n_h$ &$n_m$ & Accuracy    \\
    \midrule
    LSTM              & 41   & -    & 55.9 $\pm$ 5.3 \\
    LMN               & 52   & -    & 55.0 $\pm$1.0  \\
    CW-RNN            & 13   & -    & 74.4 $\pm$2.9 \\
    MS-LMN            & 25   & 25   & 78.0 $\pm$3.4   \\
    pret-MS-LMN       & 25   & 25   & \textbf{79.6 $\pm$3.8}   \\
    \bottomrule
    \end{tabular}
\end{table}
\subsection{Handwritten Character Recognition}
IAM-OnDB~\cite{gravesUnconstrainedOnlineHandwriting2008} is a dataset of handwritten digits. The dataset is composed of $12,179$ handwritten sentences. Each sentence is represented as a sequence of pen strokes, where each stroke is a sequence of temporal positions of the pen on the whiteboard. Each measurement is represented as a quadruple \mbox{$<time,\ x,\ y,$} $end\_stroke>$, where $time$ is the timestamp of the measurement, $x$ and $y$ the coordinates of the pen on the whiteboard, and $end\_stroke$ is $1$ for the last measurement of each stroke and $0$ otherwise. The labels are the corresponding character, lower case or upper case, digits, and some special characters. We extracted a different sample from each line in the dataset and we split the data into a train, validation and test set, of $5,364$, $3,859$, and $2,956$ sentences respectively. We normalized each stroke separately such that the timestamp and positions lie in the range $[0, 1]$. Given a sequence of pen measurements, the model must predict the corresponding sequence of characters.

A single layer bidirectional network is trained with connectionist temporal classification (CTC) loss~\cite{gravesConnectionistTemporalClassification} for each model. In order to perform a fair comparison, all models use a number of hidden units corresponding to approximately $100,000$ parameters, with $9$ memory modules for the CW-RNN and MS-LMN. For the incremental training, we add a new module every $50$ epochs. We stop the training once the validation loss does not improve for $50$ epochs. In our preliminary experiments all the models were highly sensitive to the random seed and initialization. Therefore, we decided to train each model $5$ times and report the test set results of the best model for each configuration (selected on the validation set).

Table \ref{tbl:IAM-OnDB} shows the results of the experiment.
On this dataset, the incremental training is fundamental to achieve a good performance on the Multiscale LMN, which allows to double the accuracy of the model. However, the best accuracy is obtained by the LSTM. We hypothesize that the use of gates in a multiscale architecture like the CW-RNN and the MS-LMN could further improve the performance of these models. The result also suggests that the LSTM's gates are effectively adding expressive power to the model in addition to their role in reducing the vanishing gradient since the multiscale architecture is already able to reduce the vanishing gradient. While the maximum number of modules for the hierarchical models is set to $9$, the incremental training stops at $8$ modules due to the early stopping criterion. This result shows another advantage of the incremental training, which is the ability to adaptively select the model complexity.
\begin{table}[t]
    \centering
    \caption{Best path accuracy computed on IAM-OnDB online test set.}\label{tbl:IAM-OnDB}
    \begin{tabular}{lllll}
    \toprule
           & $n_h$ & $n_m$ & Accuracy \\
    \midrule
    LSTM   & $110$ &       & \textbf{77.3}     \\
    CW-RNN & $33$  &       & 32.9     \\
    MS-LMN & $144$ & $16$  & 26.5     \\
    pret-MS-LMN & $144$ & $16$  & 66.8 \\
    \bottomrule
    \end{tabular}
\end{table}

\section{Related Work}

Recurrent neural networks such as the vanilla RNN~\cite{elmanFindingStructureTime1990} and the LSTM~\cite{hochreiterLongShortTermMemory1997} are state-of-the-art models on many sequence classification problems, for example speech recognition~\cite{gravesSPEECHRECOGNITIONDEEP2013} or handwritten character recognition~\cite{UnconstrainedOnlineHandwriting}. However, due to the vanishing gradient problem~\cite{pascanuDifficultyTrainingRecurrent2013,hochreiterVanishingGradientProblem1998}, it is difficult to learn long-term dependencies with these models. Hierarchical RNNs, such as the Clockwork RNN~\cite{koutnikClockworkRNN2014}, Phased LSTM~\cite{neilPhasedLSTMAccelerating2016}, and Hierarchical Multiscale RNN~\cite{chungHIERARCHICALMULTISCALERECURRENT2017}, solve this limitation by modifying the architecture to easily encode long-term dependencies in the hidden state. Most of the effort of the literature focus on architectural modifications~\cite{chungHIERARCHICALMULTISCALERECURRENT2017,Mali2019TheNS,neilPhasedLSTMAccelerating2016}. Another line of research explores the use of online algorithms to train RNNs~\cite{ororbiaContinualLearningRecurrent2020,NIPS2018_7991,tallecUnbiasedOnlineRecurrent2018}. Instead, in this paper, we additionally propose an incremental training algorithm that exploits the modularity of the hierarchical architecture during training. 

The MS-LMN is based on the Linear Memory Network~\cite{bacciuLinearMemoryNetworks2019}, and the initialization procedure exploits the training algorithm for the LAES~\cite{sperdutiEfficientComputationRecursive2007}. 
While \cite{bacciuLinearMemoryNetworks2019} proposes a pretraining algorithm for the LMN that also exploits the LAES, the procedure proposed in this paper for the MS-LMN represents an improvement under several aspects. First, it does not require the expensive pretraining of the unrolled model. Furthermore, the incremental approach is less computationally expensive for long sequences.

The incremental training can be seen as a form of curriculum training~\cite{bengioCurriculumLearning2009} since the model is trained on gradually more difficult problems by increasing the complexity of the network. The main difference is that traditionally in the curriculum learning scenario the model is fixed while the data is changing, while in the incremental training it is the model which is changed to model the more difficult long-term dependencies.

\section{Conclusion}
Time series datasets with high-frequency samples provide challenging environments for learning with RNNs. The MS-LMN provides a natural solution for the problem by separating the memory state of the model into several modules, each one updated at different frequencies. Furthermore, the incremental training algorithm helps to learn long-term dependencies by incrementally adding new modules to the network. The experimental results show that the modular network can effectively learn features at different frequencies, such as modelling low-frequency and high-frequency changes in a generated sequence. Furthermore, the incremental training provides a consistent improvement to the final performance compared to traditional stochastic gradient descent.

In the future, we plan to apply the incremental training algorithm to novel architectures and learning settings and to optimize it to improve the convergence speed, for example by considering more sophisticated criteria for the dynamic addition of a module. The modular architecture and incremental learning can be also exploited in different learning settings, such as multi-task or continual learning scenarios, where the modular separation can be evinced from the data.

\section*{Acknowledgments}
This work has been supported by MIUR under project SIR 2014 LIST-IT (RBSI14STDE) and by the DEEPer project, University of Padova.

%% file: img/lmn.tex
\definecolor{blue_light}{RGB}{218,232,252}
\definecolor{blue_dark}{RGB}{108,142,191}
\definecolor{green_light}{RGB}{213,232,212}
\definecolor{green_dark}{RGB}{130,179,102}
\definecolor{yellow_light}{RGB}{255,242,204}
\definecolor{yellow_dark}{RGB}{214,182,86}

\begin{tikzpicture} 
 
    \tikzstyle{sty_memory} = [draw=blue_dark,fill=blue_light,circle,outer sep=0,inner sep=1,minimum size=25]
    \tikzstyle{sty_functional} = [draw=green_dark,fill=green_light,circle,outer sep=0,inner sep=1,minimum size=25]
    \tikzstyle{sty_out} = [draw=yellow_dark,fill=yellow_light,circle,outer sep=0,inner sep=1,minimum size=25]
    \tikzstyle{myedgestyle} = [-latex]
    \tikzstyle{sty_module} = [rounded corners,fill opacity=0.4]

    \draw [draw=green_dark,fill=green_light,sty_module] (-2.25,-2) rectangle (-0.25,1.75);   
    \node[sty_functional] (x) at (-1.25, -1) {$\vx^t$};
    \node[sty_functional] (h) at (-1.25, 1) {$\vh^t$};
    \draw [myedgestyle] (x) edge (h);
    \node[sty_out] (y) at (1, 3) {$\vy^t$};

    \draw [draw=blue_dark,fill=blue_light,sty_module] (0.25,-2) rectangle (1.75,1.75);
    \node[sty_memory] (m) at (1, 1) {$\vm^t$};
    \node[sty_memory] (mp) at (1, -1) {$\vm^{t-1}$};
    \draw [myedgestyle] (h) edge (m);
    \draw [myedgestyle] (mp) edge (m);
    \draw [myedgestyle] (mp) edge (h);
    \draw [myedgestyle] (m) edge (y);
    
    \draw[very thin] (1.5, 1) -- (2, 1) -- (2, -1);
    \draw[very thin, ->] (2, -1) -- (1.5, -1);
    \fill [black] (1.9,-0.1) rectangle (2.1,0.1);

    \node[] (f_text) at (-1.25, -1.75) {$Functional$};
    \node[] (f_text) at (1, -1.75) {$Memory$};
\end{tikzpicture}

%% file: img/ms_lmn.tex
\definecolor{blue_light}{RGB}{218,232,252}
\definecolor{blue_dark}{RGB}{108,142,191}
\definecolor{green_light}{RGB}{213,232,212}
\definecolor{green_dark}{RGB}{130,179,102}
\definecolor{yellow_light}{RGB}{255,242,204}
\definecolor{yellow_dark}{RGB}{214,182,86}

\begin{tikzpicture}

    \tikzstyle{sty_memory} = [draw=blue_dark,fill=blue_light,circle,outer sep=0,inner sep=1,minimum size=25]
    \tikzstyle{sty_functional} = [draw=green_dark,fill=green_light,circle,outer sep=0,inner sep=1,minimum size=25]
    \tikzstyle{sty_out} = [draw=yellow_dark,fill=yellow_light,circle,outer sep=0,inner sep=1,minimum size=25]
    \tikzstyle{myedgestyle} = [-latex]
    \tikzstyle{sty_module} = [rounded corners,fill opacity=0.1]
    \tikzstyle{sty_module_all} = [rounded corners,fill opacity=0.4]

    \draw [draw=green_dark,fill=green_light,sty_module_all] (-2.25,-2) rectangle (-0.25,1.75);   
    \node[sty_functional] (x) at (-1.25, -1) {$\vx^t$};
    \node[sty_functional] (h) at (-1.25, 1) {$\vh^t$};
    \draw [myedgestyle] (x) edge (h);
    \node[sty_out] (y) at (3, 3) {$\vy^t$};

    \draw [draw=blue_dark,fill=blue_light,sty_module_all] (0.,-2.5) rectangle (6,2);
    
    \draw [draw=blue_dark,fill=blue_light,sty_module] (0.25,-2) rectangle (1.75,1.75);
    \node[sty_memory] (m) at (1, 1) {$\vm_1^t$};
    \node[sty_memory] (mp) at (1, -1) {$\vm_1^{t-1}$};
    \draw [myedgestyle, dashed] (h) edge (m);
    \draw [myedgestyle, dashed] (mp) edge (m);
    \draw [myedgestyle] (mp) edge (h);
    \draw [myedgestyle] (m) edge (y);
    \node at (1.0, -1.75) {$f_1=1$};

    \draw [draw=blue_dark,fill=blue_light,sty_module] (2.25,-2) rectangle (3.75,1.75);
    \node[sty_memory] (m2) at (3, 1) {$\vm_2^t$};
    \node[sty_memory] (mp2) at (3, -1) {$\vm_2^{t-1}$};
    \draw [myedgestyle, bend left, dashed] (h) edge (m2);
    \draw [myedgestyle, dashed] (mp2) edge (m2);
    \draw [myedgestyle] (mp2) edge (h);
    \draw [myedgestyle] (m2) edge (y);
    \draw [myedgestyle, dashed] (mp2) edge (m);
    \node at (3.0, -1.75) {$f_2=2$};

    \draw [draw=blue_dark,fill=blue_light,sty_module] (4.25,-2) rectangle (5.75,1.75);
    \node[sty_memory] (m3) at (5, 1) {$\vm_3^t$};
    \node[sty_memory] (mp3) at (5, -1) {$\vm_3^{t-1}$};
    \draw [myedgestyle, bend left, dashed] (h) edge (m3);
    \draw [myedgestyle, dashed] (mp3) edge (m3);
    \draw [myedgestyle] (mp3) edge (h);
    \draw [myedgestyle] (m3) edge (y);
    \draw [myedgestyle, dashed] (mp3) edge (m);
    \draw [myedgestyle, dashed] (mp3) edge (m2);
    \node at (5.0, -1.75) {$f_3=4$};
    
    \draw[very thin] (1.5, 1) -- (2, 1) -- (2, -1);
    \draw[very thin, ->] (2, -1) -- (1.5, -1);
    \fill [black] (1.9,-0.5) rectangle (2.1,-0.3);
    
    \draw[very thin] (3.5, 1) -- (4, 1) -- (4, -1);
    \draw[very thin, ->] (4, -1) -- (3.5, -1);
    \fill [black] (3.9,-0.5) rectangle (4.1,-0.3);
    
    \draw[very thin] (5.5, 1) -- (5.9, 1) -- (5.9, -1);
    \draw[very thin, ->] (5.9, -1) -- (5.5, -1);
    \fill [black] (5.8,-0.5) rectangle (6.0,-0.3);

    \node[] (f_text) at (-1.25, -1.75) {$Functional$};
    \node[] (f_text) at (3, -2.25) {$Modular\ Memory$};
\end{tikzpicture}

%% file: img/mslmn_blocks.tex
\begin{figure}[tb]
    \centering
    \resizebox{0.75\linewidth}{!}{
    \begin{tikzpicture}[
        textnode/.style={},
        squareactive/.style={rectangle, draw=black, fill=black!25, minimum size=5mm},
        squareinactive/.style={rectangle, draw=black, fill=black!5, minimum size=5mm},
        squareempty/.style={rectangle, draw=black!0, fill=black!0, minimum size=5mm},
        square/.style={rectangle, draw=black, fill=black!0, minimum size=5mm}
    ]
    
    \begin{scope}[local bounding box=ht]
    \foreach \i in {0,...,4} {
        \ifthenelse{\i < 3} {
            \node[squareinactive, minimum width=0.3cm] (ht\i) at (0,0.5*\i) {};
        }{
            \node[squareactive, minimum width=0.3cm] (ht\i) at (0,0.5*\i) {};
        }
    }
    \end{scope}
    
    \begin{scope}[shift={($(ht.east)+(2.5cm,-1cm)$)}, local bounding box=Wxh]
    \foreach \i in {0,...,4} {
        \ifthenelse{\i < 3} {
            \node[squareinactive, minimum width=1.0cm] (Wxh\i) at (0,0.5*\i) {};
        } {
            \node[squareactive, minimum width=1.0cm] (Wxh\i) at (0,0.5*\i) {};
        }
    }
    \end{scope}
    
    \begin{scope}[shift={($(Wxh.east)+(0.7cm,-1cm)$)}, local bounding box=xt]
    \node[squareempty, minimum width=0.3cm] (xt0) at (0,0) {};
    \node[squareempty, minimum width=0.3cm] (xt1) at (0,1.75cm) {};
    \node[squareactive, minimum width=0.3cm, minimum height=1.0cm] (xt) at (0,1.0cm) {};
    \end{scope}
    
    \begin{scope}[shift={($(xt.east)+(2cm,-1cm)$)}, local bounding box=Whh]
    \node[square, minimum size=2.5cm] (rett) at (1,1) {};
    \foreach \i in {0,...,4} {
        \foreach \j in {0,...,4} {
            \pgfmathtruncatemacro{\invj}{4 - \j};
            \ifthenelse{\i < \invj} {
            }{
                \ifthenelse{\invj < 2} {
                    \node[squareactive] (Whh\i\j) at (0.5*\i,0.5*\j) {};
                } {
                    \node[squareinactive] (Whh\i\j) at (0.5*\i,0.5*\j) {};
                }
            }
        }
    }
    \node[textnode] (Whhzero) at (0.25,0.25) {\LARGE\( 0 \)};
    \end{scope}
    
    \begin{scope}[shift={($(Whh.east)+(0.7cm,-1cm)$)}, local bounding box=hprev]
    \foreach \i in {0,...,4} {
        \node[squareactive, minimum width=0.3cm] (hprev\i) at (0,0.5*\i) {};
    }
    \end{scope}

    
    \node[textnode] (textht) at ($(ht.south)+(0,-0.7cm)$) {\( \vm^t \)};
    \node[textnode] (textsigma) at ($(Wxh.south west)+(-0.55cm,-0.7cm)$) {\( \sigma \) \Large \( ( \)};
    \node[textnode] (textequals) at ($(textht.east)!0.5!(textsigma.west)+(0,-0.1cm)$) {\( = \)};
    \node[textnode] (textWxh) at ($(Wxh.south)+(0,-0.7cm)$) {\( \mW^{hm} \)};
    \node[textnode] (textxt) at ($(xt.south)+(0,-1.46cm)$) {\( \vh^{t} \)};
    \node[textnode] (textWhh) at ($(Whh.south)+(0,-0.7cm)$) {\( \mW^{mm} \)};
    \node[textnode] (textplus) at ($(textxt.east)!0.5!(textWhh.west)+(0,-0.05cm)$) {\( + \)};
    \node[textnode] (texthprev) at ($(hprev.south)+(0,-0.7cm)$) {\( \vm^{t-1} \)};
    \node[textnode] (textClose) at ($(hprev.south east)+(0.4,-0.7cm)$) {\Large\( ) \)};
    
    \foreach \i in {0,...,4} {
        \pgfmathtruncatemacro{\invi}{5 - \i};
        \node[textnode] (texthtm\i) at ($(ht.south east)+(0.35,0.25+0.5*\i)$) {\scriptsize\( h_{\invi} \)};
    }
    
    
    \node[textnode] (textht1) at ($(ht.west)+(-0.4,0)$) {\footnotesize\( g N_m \)};
    \node[textnode] (textWh1) at ($(Whh.west)+(-0.4,0)$) {\footnotesize\( g N_m \)};
    \node[textnode] (textWh2) at ($(Whh.north)+(0,0.25)$) {\footnotesize\( g N_m \)};
    \node[textnode] (textxprev1) at ($(hprev.east)+(+0.45,0)$) {\footnotesize\( g N_m \)};
    \node[textnode] (textWx1) at ($(Wxh.west)+(-0.4,0)$) {\footnotesize\( g N_m \)};
    \node[textnode] (textWx2) at ($(Wxh.north)+(0,0.25)$) {\footnotesize\( N_h \)};
    \node[textnode] (textxt1) at ($(xt.east)+(+0.35,0)$) {\footnotesize\( N_h \)};
    
    \end{tikzpicture}
    }
        
    \caption{Representation of the memory update with block matrices showing the size for \( g = 5 \), assuming that only the first two modules are active at time \( t \). Darker blocks represent the active weights.}\label{fig:cwrnn-blocks}
\end{figure}

%% file: img/ms_lmn_init.tex
\definecolor{blue_light}{RGB}{218,232,252}
\definecolor{blue_dark}{RGB}{108,142,191}
\definecolor{green_light}{RGB}{213,232,212}
\definecolor{green_dark}{RGB}{130,179,102}
\definecolor{yellow_light}{RGB}{255,242,204}
\definecolor{yellow_dark}{RGB}{214,182,86}
\definecolor{brick_red}{rgb}{0.8, 0.0, 0.0}

\begin{tikzpicture}
    \tikzstyle{sty_memory} = [draw=blue_dark,fill=blue_light,circle,outer sep=0,inner sep=1,minimum size=25]
    \tikzstyle{sty_functional} = [draw=green_dark,fill=green_light,circle,outer sep=0,inner sep=1,minimum size=25]
    \tikzstyle{sty_out} = [draw=yellow_dark,fill=yellow_light,circle,outer sep=0,inner sep=1,minimum size=25]
    \tikzstyle{myedgestyle} = [-latex]
    \tikzstyle{sty_newedge} = [line width=1pt,draw=brick_red,-latex]
    \tikzstyle{sty_module} = [rounded corners,fill opacity=0.4]

    \draw [draw=black,sty_module] (-2,-2.25) rectangle (2,2);
    \node[] (f_text) at (0, -2) {Trained Model};
    \draw [draw=black,sty_module] (2,-2.25) rectangle (4,2);
    \node[] (f_text) at (3, -2) {New Module};

    \draw [draw=green_dark,fill=green_light,sty_module] (-1.75,-1.75) rectangle (-0.25,1.75);   
    \node[sty_functional] (x) at (-1, -1) {$\vx^t$};
    \node[sty_functional] (h) at (-1, 1) {$\vh^t$};
    \draw [myedgestyle] (x) edge (h);
    \node[sty_out] (y) at (1, 3) {$\vy^t$};

    \draw [draw=blue_dark,fill=blue_light,sty_module] (0.25,-1.75) rectangle (1.75,1.75);
    \node[sty_memory] (m) at (1, 1) {$\vm_1^t$};
    \node[sty_memory] (mp) at (1, -1) {$\vm_1^{t-1}$};
    \draw [myedgestyle, dashed] (h) edge (m);
    \draw [myedgestyle, dashed] (mp) edge (m);
    \draw [myedgestyle] (mp) edge (h);
    \draw [myedgestyle] (m) edge (y);

    \draw [dashed, draw=blue_dark,fill=blue_light,sty_module] (2.25,-1.75) rectangle (3.75,1.75);
    \node[dashed, sty_memory] (m2) at (3, 1) {$\vm_2^t$};
    \node[dashed, sty_memory] (mp2) at (3, -1) {$\vm_2^{t-1}$};
    \draw [sty_newedge, bend left, dashed] (h) edge node[above=1cm, left=.7cm of mp, text=brick_red] {$\mA_{g+1}$} (m2);
    \draw [sty_newedge, dashed] (mp2) edge node[right, text=brick_red] {$\mB_{g+1}$} (m2);
    \draw [sty_newedge, dashed] (mp2) edge (h);
    \draw [sty_newedge] (m2) edge node[above, right=0.2cm of y, text=brick_red] {$\mW^{out}_{g+1}$} (y);
    \draw [sty_newedge, dashed] (mp2) edge (m);
    
    \draw[very thin] (1.5, 1) -- (1.9, 1) -- (1.9, -1);
    \draw[very thin, ->] (1.9, -1) -- (1.5, -1);
    \fill [black] (1.8,-0.5) rectangle (2.,-0.3);
    
    \draw[very thin] (3.5, 1) -- (3.9, 1) -- (3.9, -1);
    \draw[very thin, ->] (3.9, -1) -- (3.5, -1);
    \fill [black] (3.8,-0.5) rectangle (4.,-0.3);
    
\end{tikzpicture}

%% file: pcode/pseudo_train_ms_lmn.tex
\begin{algorithm}
    \caption{MS-LMN training}\label{alg:ms_lmn_train}
    \begin{algorithmic}[1]
        \Procedure{MS-LMNTrain}{$Data$, $N_h$, $N_m$, $g$}
            \For {\( i \in \{0, \hdots, g-1\} \)}
                \State \( \textit{ms-lmn.add-module}(Data, i) \)
                \State \( \textit{ms-lmn.fit}(Data) \)
            \EndFor
            \State \Return \( \textit{ms-lmn} \) 
        \EndProcedure
        
        \Procedure{add-module}{$self$, $Data$, $g$}
            \State \( \mH_g = [\ ] \)
            \For {\(seq \in Data\)}
                \State \( \vh_{seq}, \vm_{seq} \gets self(seq) \)
                \State \( \vh_{seq}.subsample(2^g)\)
                \State \( \mH_g.append(\vh_{seq}) \)
            \EndFor
            \State \( laes \gets \textit{train-laes}(N_m) \)
            \State \( laes.fit(\mH_g) \)
            \State \( self.\mW^{h m_{g+1}} \gets laes.\mA \)
            \State \( self.\mW^{m_{g+1} m_{g+1}} \gets laes.\mB \)
            \State \( self.\mW^{m_{g+1} o} \gets \textit{fit\_readout}(self,\ laes, Data) \)  \Comment{computed with the pseudoinverse}
        \EndProcedure
    \end{algorithmic}
\end{algorithm}	

%% file: img/gen_seq.tex
\begin{tikzpicture}[scale=0.7]

\node[label={RNN (NMSE=$79.5$)},draw=none,fill=none] (rnn) at (0,-1.5){\includegraphics[width=.3\textwidth]{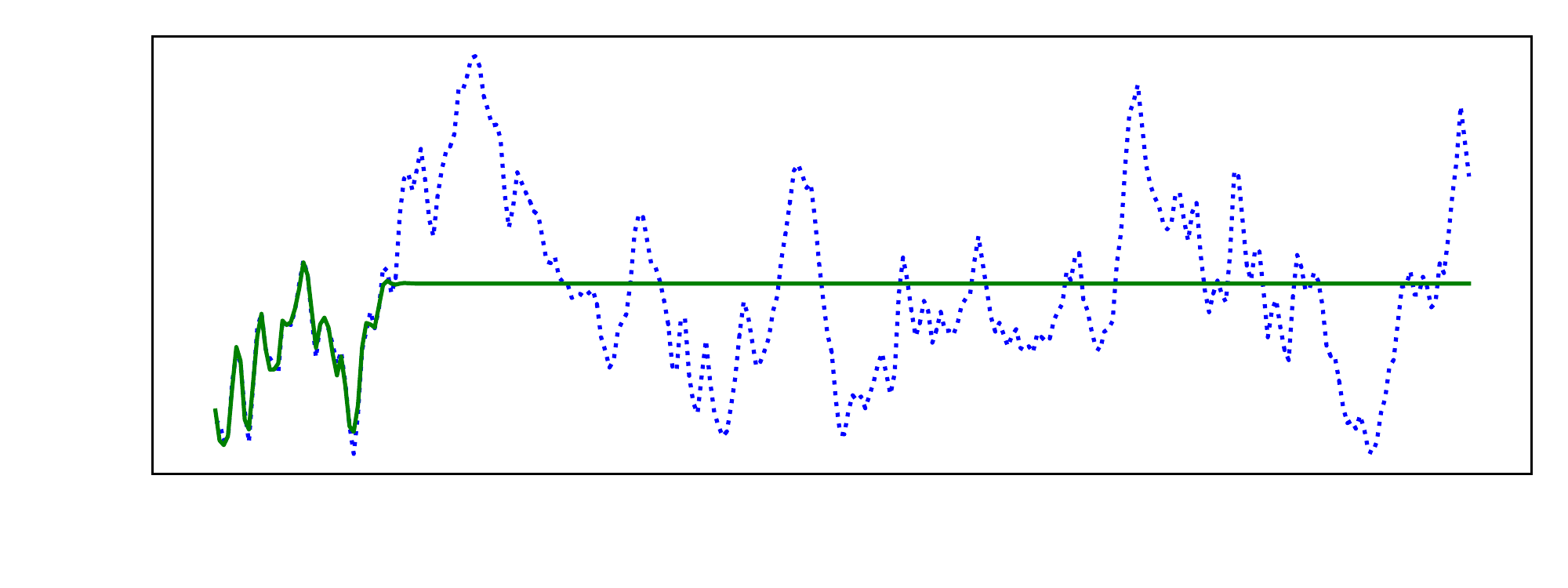}};
\node[label={LMN (NMSE=$38.4$)},draw=none,fill=none] (lmn) at (0, -4.5) {\includegraphics[width=.3\textwidth]{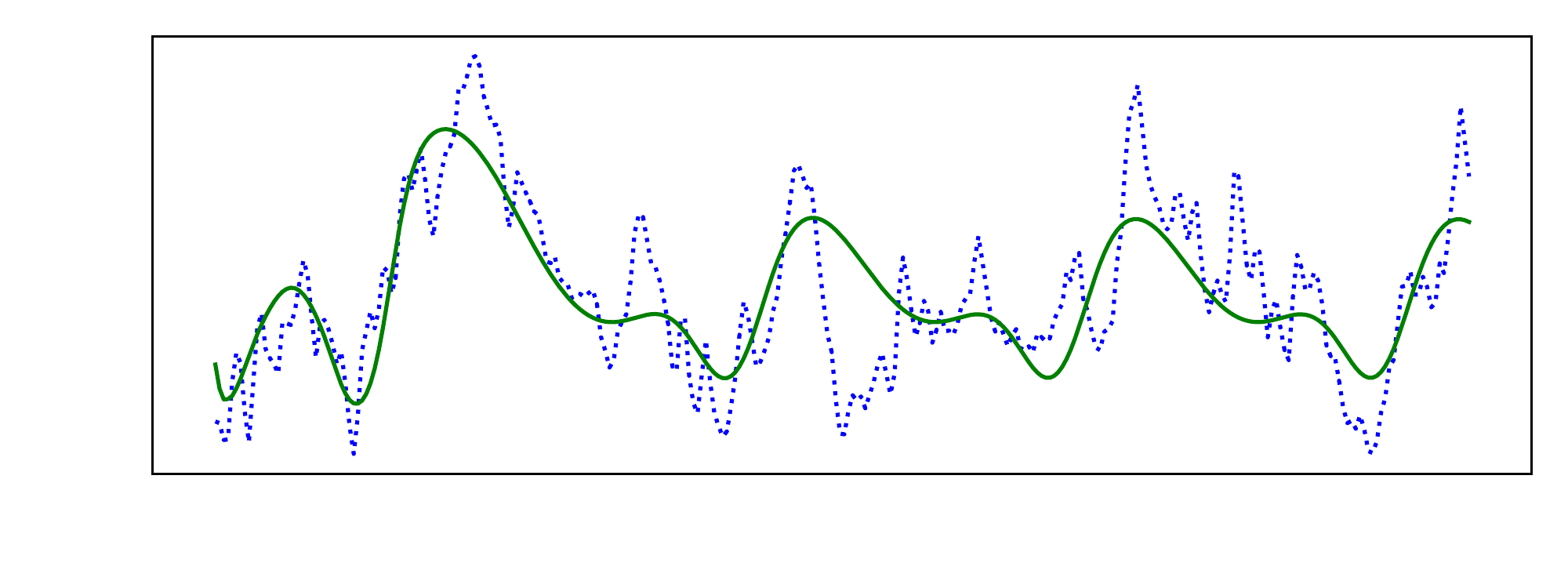}};

\node[label={CW-RNN (NMSE=$12.5$)},draw=none,fill=none] (cwrnn) at (6, -1.5) {\includegraphics[width=.3\textwidth]{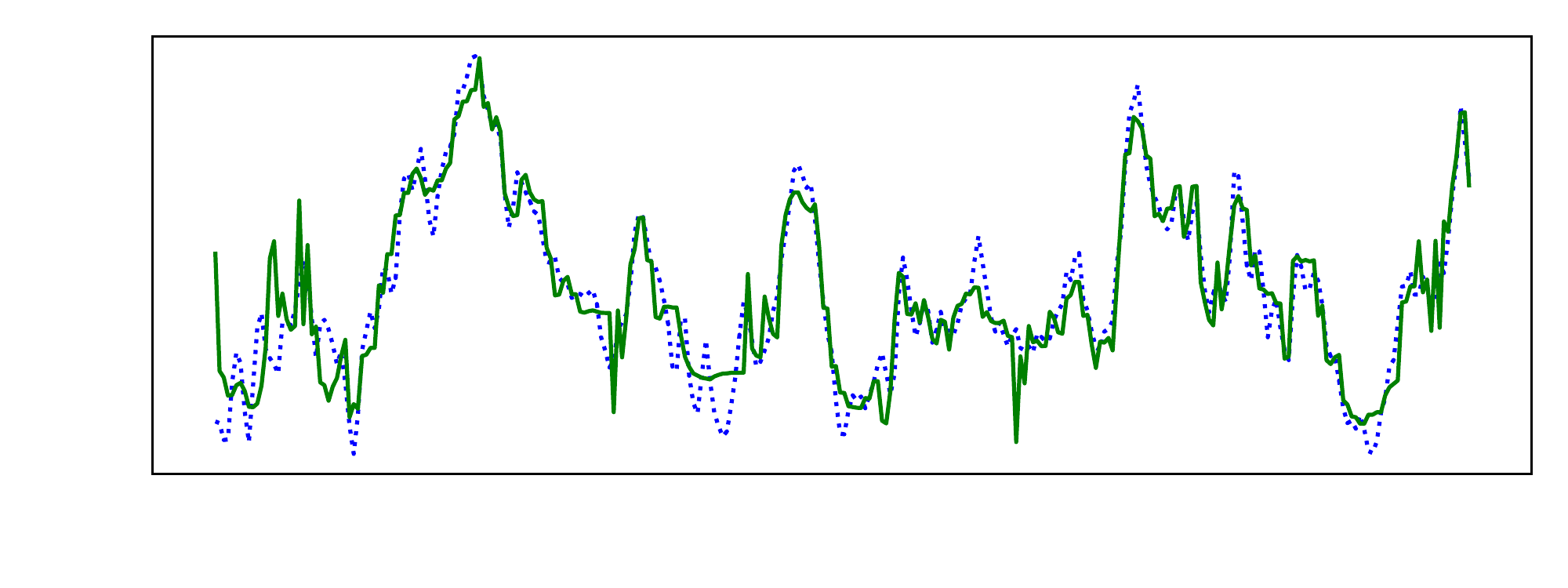}};
\node[label={MS-LMN (NMSE=$0.116$)},draw=none,fill=none] (cw-lmn) at (6, -4.5) {\includegraphics[width=.3\textwidth]{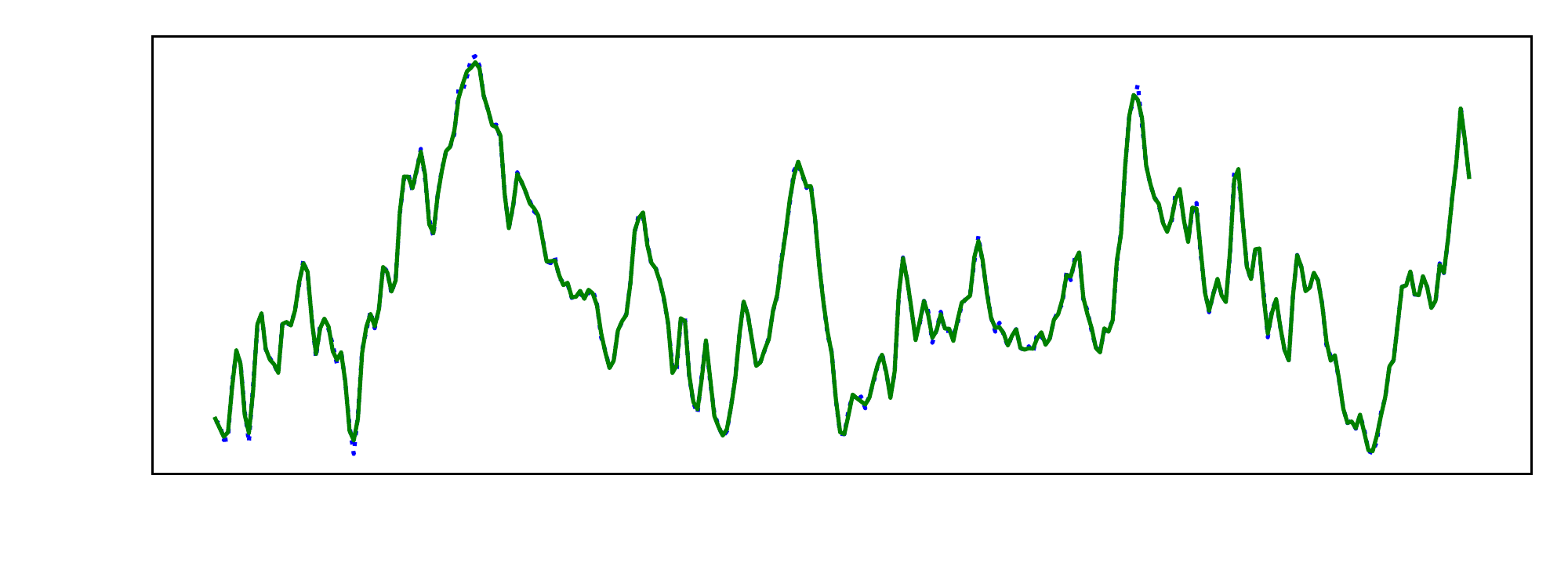}};

\end{tikzpicture}